\documentclass[sigconf,nonacm]{acmart}
\setcopyright{none}
\usepackage{booktabs}
\usepackage{multirow}
\usepackage{tabularx}
\usepackage{placeins}

\AtBeginDocument{%
  }

\begin{document}

\title{\textsc{Autoschema}: Live Schema Grounding for Agentic
Text-to-\textsc{Sparql} over Biomedical Knowledge Graphs}
\title{AutoSchema: Live Schema Grounding for Agentic
Text-to-\textsc{Sparql} over Heterogeneous Knowledge Graphs}

\author{Yiming Zhang}
\email{zhang\_yiming\_25@stu-cbms.k.u-tokyo.ac.jp}
\affiliation{%
  \institution{The University of Tokyo}
  \city{Kashiwa}
  \state{Chiba}
  \country{Japan}
}

\author{Koji Tsuda}
\email{tsuda@k.u-tokyo.ac.jp}
\affiliation{%
  \institution{Center for Basic Research on Materials, National Institute for Materials Science}
  \city{Tsukuba}
  \state{Ibaraki}
  \country{Japan}
}
\affiliation{%
  \institution{RIKEN Center for Advanced Intelligence Project}
  \city{Tokyo}
  \country{Japan}
}


\begin{abstract}
Life science knowledge graphs make large collections of structured data
available through SPARQL, but each resource uses its own schema, identifiers,
and links.  TogoMCP helps language model agents query these resources by
providing curated Metadata Interoperability Exchange files.  Creating and
maintaining these files still requires language model assisted drafting,
validation, and manual review.  We study \emph{live schema grounding}, where
an agent obtains the schema evidence needed for a question directly from the
current endpoints.  We present \textsc{autoschema}, a general framework for
live schema grounding that requires no training.  It inspects live schemas,
maps entity names in a question to graph identifiers, explores relation paths,
and finds possible
connections between resources during iterative query construction.  We use
TogoMCP as our main comparison framework.  We evaluate
\textsc{autoschema} on Resource
Focused Biomedical KGQA, Multi Resource Biomedical KGQA, Longitudinal Biomedical
Semantic QA over BioASQ Task B, and Chemistry Knowledge Graph Transfer to a
previously undocumented RDF graph.  \textsc{autoschema} improves mean factoid
accuracy over TogoMCP in the biomedical KGQA tasks and gives
consistent gains in the longitudinal BioASQ evaluation.  It also reduces
iteration budget exhaustion and uses fewer tool calls on average in the core
evaluation.  The transfer study gives preliminary evidence that live schema
grounding can support irregular and previously unseen graphs without first
creating a curated schema file.
\end{abstract}


\maketitle

\begin{figure*}[t]
  \centering
  \includegraphics[width=\textwidth]{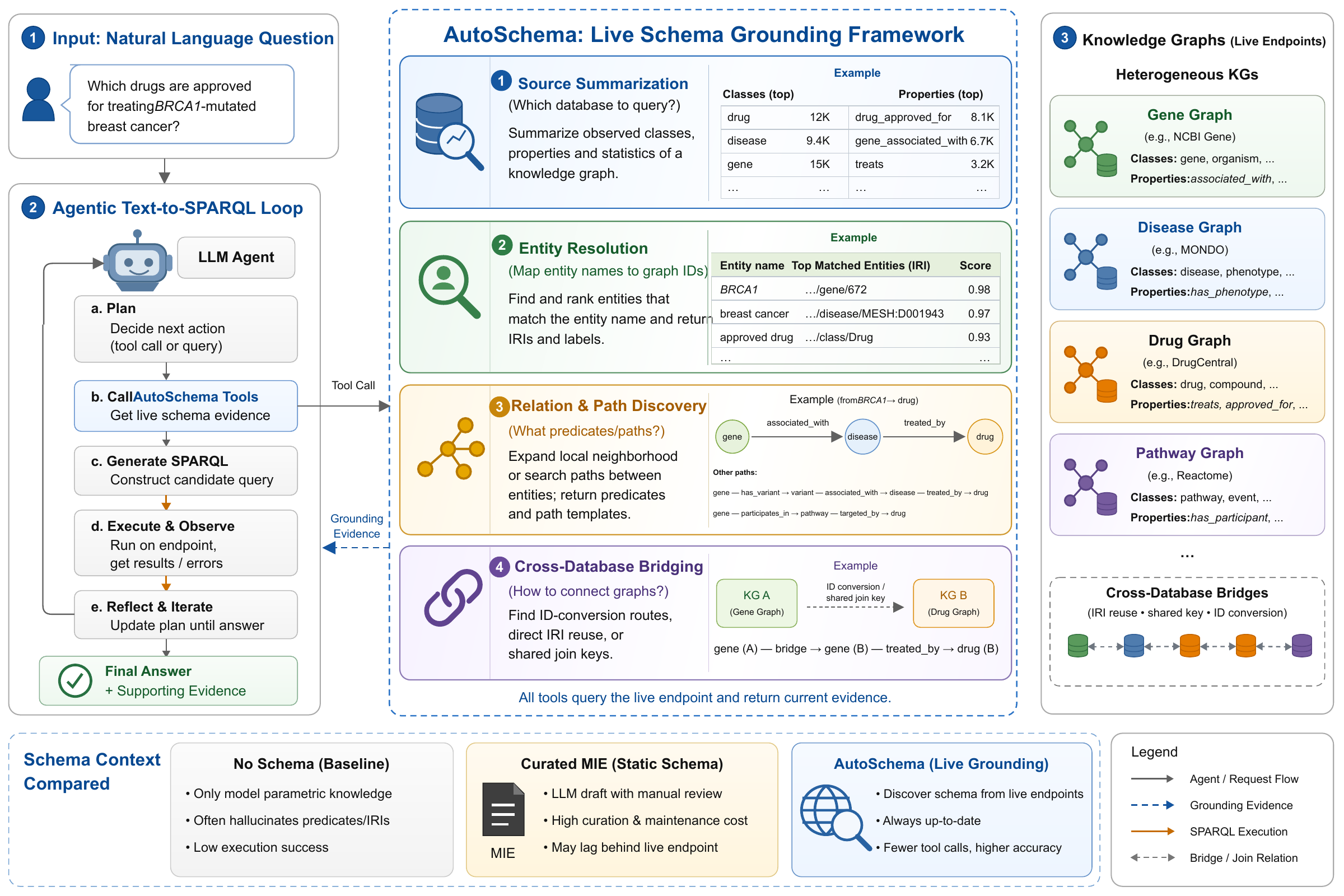}
  \caption{AutoSchema in an agentic text-to-SPARQL loop over heterogeneous
  knowledge graphs.  Given a natural-language question, the agent alternates
  planning, live grounding, query generation, execution, and reflection.
  AutoSchema supplies four question-time capabilities: source summarization,
  entity resolution, relation and path discovery, and cross-database bridging.
  Each capability returns current endpoint evidence rather than a complete
  schema dump.  The bottom panel contrasts AutoSchema with TogoMCP
  and a no-schema lower bound.}
  \Description{The left column shows a natural-language question and an LLM
  agent iterating through planning, AutoSchema tool calls, SPARQL generation,
  endpoint execution, and reflection before producing an answer.  The center
  column details four live grounding capabilities with examples: source
  summarization, entity resolution, relation and path discovery, and
  cross-database bridging.  The right column shows heterogeneous gene, disease,
  drug, and pathway graphs connected by possible bridges.  A bottom comparison
  summarizes the no-schema lower bound, TogoMCP, and AutoSchema.}
  \label{fig:autoschema-overview}
\end{figure*}

\section{Introduction}
\label{sec:introduction}

Life science knowledge is spread across specialized resources for genes,
proteins, variants, pathways, diseases, and compounds.  RDF gives these
resources a common data model, and SPARQL supports precise queries over their
structured content~\cite{callahan2013bio2rdf,sima2019semantic}.  The RDF Portal
brings together many reviewed life science knowledge graphs and provides
public SPARQL endpoints~\cite{kawashima2018nbdc}.  These resources support data
integration and biomedical discovery~\cite{kamdar2019webscale}.  However, a
researcher still needs detailed technical knowledge to use them.  Answering a
natural language question requires choosing a resource, mapping entity names
to database identifiers, selecting valid predicates, and following links
between resources.  Natural language interfaces can lower this barrier, but
database specific schemas remain a central challenge
~\cite{marginean2017gfmed,sima2022biosoda}.

Language models offer a flexible way to translate questions into SPARQL.
Without schema evidence, they can generate valid looking queries that use
predicates or identifiers that are absent from the target graph
~\cite{dabramo2025investigating,mecharnia2025performance}.  Existing methods
reduce these errors by training on paired questions and queries, retrieving
examples, or adding schema descriptions to the model context
~\cite{li2023kbbinder,luo2024chatkbqa,gao2025schema,
emonet2024expasy,nagazumi2026accurate}.  These methods show that schema
grounding helps, but they usually assume that useful schema material already
exists.

TogoMCP~\cite{kinjo2026togomcp} is the closest starting point for our work.  It gives an agent access
to the RDF Portal through Model Context Protocol tools.  It also provides a
Metadata Interoperability Exchange (MIE) file for each supported resource.  An MIE
file contains schema descriptions, working SPARQL examples, common errors,
and cross resource query patterns. TogoMCP shows that this context is a main
source of its performance gain~\cite{kinjo2026togomcp}.  We therefore use
TogoMCP as our main baseline, rather than treating MIE as a separate
system.

MIE creation is a mixed automatic and manual process.  The reported TogoMCP workflow uses
automatic schema discovery, language model assisted drafting, endpoint
validation, and manual review.  Its reported implementation uses hosted
Claude models, although the MIE format itself is not tied to one model family
~\cite{kinjo2026togomcp}.  This workflow creates two practical limits.  First,
private RDF data may not be allowed to leave local infrastructure through a
hosted model API.  Second, exploring an endpoint and drafting a long MIE file
can require many paid model tokens.  A curated file must also be checked and
updated when its endpoint changes.  These costs slow down the addition of new
resources and can leave an existing file out of date.

We ask whether an agent can build executable SPARQL by discovering the needed
schema directly from live endpoints at question time.  We call this problem
\emph{live schema grounding}.  It covers source selection, entity resolution,
relation and path selection, and links between resources.  Solving this
problem would let an agent use a new or updated graph without first creating
an MIE file.  It would also keep each query decision tied to current endpoint
evidence.  When the graph is private, the complete workflow could remain on
local infrastructure.

We introduce \textsc{autoschema}, a general framework for live schema
grounding in tool using agents.  It inspects observed classes and
properties, maps entity names in a question to graph identifiers, explores
local relation paths, and finds possible links between resources.  The agent
requests this evidence only when it is needed.  It then writes a SPARQL query,
executes it, and uses the result to guide the next step.
Figure~\ref{fig:autoschema-overview} shows this process.  In our controlled
evaluation, we compare \textsc{autoschema} with TogoMCP and with a
no-schema lower bound.  We run the agents with models that have open weights
and are hosted on local infrastructure.
Schema discovery and question answering therefore do not require a closed
model API.

Our evaluation has three levels.  At the graph query level, \emph{Resource
Focused Biomedical KGQA} tests questions answered from one RDF resource, and
\emph{Multi Resource Biomedical KGQA} tests questions that combine evidence or
identifiers from several resources.  At the general biomedical question
answering level, \emph{Longitudinal Biomedical Semantic QA} uses factoid, list,
and yes or no questions from yearly BioASQ Task B test sets
~\cite{krithara2023bioasq}.  These questions also test whether the agent can
decide when graph access is useful.  At the transfer level, \emph{Chemistry
Knowledge Graph Transfer} tests a previously undocumented tmQM RDF graph with
an irregular schema~\cite{cibinel2026tmqmrdf}.

Across models with open weights, \textsc{autoschema} improves mean factoid
accuracy over TogoMCP in the biomedical KGQA tasks.  It also gives
consistent factoid gains in the longitudinal BioASQ evaluation, lowers the
rate of iteration budget exhaustion, and uses fewer tool calls on average in
the core evaluation.  The chemistry transfer result is preliminary, but it
suggests that live grounding is useful when a new graph has an irregular
schema.

Our contributions are as follows:
\begin{itemize}
  \item We formulate \emph{live schema grounding} for agentic
  text-to-SPARQL.  The task grounds query decisions in current endpoint
  evidence instead of a prewritten schema file.
  \item We develop \textsc{autoschema}, a general framework for live schema
  grounding.  It provides reusable functions for source selection, entity
  resolution, relation path discovery, and cross-resource bridging.  It does
  not require model training and can work with locally hosted models that have
  open weights.
  \item We provide a controlled evaluation across biomedical graph queries,
  longitudinal Biomedical Semantic QA, and chemistry graph transfer.  We also
  measure interaction cost and study the value of repeated schema access.
\end{itemize}

\section{Related Work}
\label{sec:related-work}

\subsection{Text-to-SPARQL and KGQA Evaluation}

Text-to-SPARQL maps a natural language question to an executable graph query.
Surveys identify entity linking, relation selection, and compositional
reasoning as core challenges~\cite{lan2021survey,chakraborty2021neural}.
Benchmarks such as WebQuestionsSP, ComplexWebQuestions, LC-QuAD~2.0, CFQ, and
KQA Pro test different forms of these challenges
\cite{yih2016webqsp,talmor2018complexweb,dubey2019lcquad,keysers2020cfq,
cao2022kqapro}.  However, evaluation protocols are often hard to compare, and
random splits can overestimate transfer to unseen relations and compositions
\cite{perevalov2022leaderboard,jiang2022generalizability}.  A system may also
receive a question whose answer is absent from the target graph
\cite{patidar2023answerability,faldu2024retinaqa}.  We therefore measure
executed answers on live endpoints and separate graph supported questions
from open biomedical questions.

\subsection{Schema Grounding and Agentic Querying}

Recent KGQA systems retrieve graph context before or during query generation.
RNG-KBQA ranks candidate queries, TIARA combines entity and schema retrieval,
and ArcaneQA limits program actions using the current graph context
\cite{ye2022rng,shu2022tiara,gu2022arcaneqa}.  KB-BINDER and ChatKBQA first
generate a logical form and then bind it to the graph
\cite{li2023kbbinder,luo2024chatkbqa}.  SG-KBQA further shows that schema
context helps with unseen graph elements~\cite{gao2025schema}.  These methods
usually assume a known target graph.  Our setting also requires the agent to
select a source and discover its current entity and relation structure.

Tool based systems make grounding iterative.  StructGPT provides interfaces
to structured data, Interactive-KBQA uses graph feedback, Think-on-Graph
traverses explicit reasoning paths, and SAGA constrains predicate exploration
with schema evidence
\cite{jiang2023structgpt,xiong2024interactive,sun2024thinkongraph,
zhang2026saga}.  Agentic SPARQL also studies endpoint discovery and federated
query formulation through MCP~\cite{dobriy2026agenticsparql}.  Our controlled
comparison keeps the agent and execution tools fixed and changes only the
source of grounding evidence.

RDF summaries offer another source of grounding evidence.  VoID describes
datasets and their links, while SchemEX, LODStats, Loupe, and ABSTAT summarize
graph structure for search or inspection
\cite{alexander2009void,konrath2012schemex,auer2012lodstats,
mihindukulasooriya2015loupe,spahiu2016abstat}.  Federated engines such as FedX
and ANAPSID instead focus on source selection and query execution
\cite{schwarte2011fedx,acosta2011anapsid}.  \textsc{Autoschema} does not replace
a federated query engine.  It turns live, question specific observations into
grounding evidence for an agent.

\subsection{Biomedical Knowledge Access and Benchmark Construction}

Biomedical Linked Data provides a demanding instance of this problem because
resources use different schemas and identifiers.  Bio2RDF introduced common
linked data conventions, BioFed supports queries across life science
endpoints, and Bio-SODA provides interactive question answering
\cite{callahan2013bio2rdf,saleem2017biofed,sima2022biosoda}.  The RDF Portal
provides reviewed datasets and public SPARQL endpoints
\cite{kawashima2018nbdc}.  RDF-config places expert schemas behind a fixed
query builder~\cite{nagazumi2026accurate}.  TogoMCP instead gives LLM agents
direct access to the endpoints and supplies a compact MIE file for each
resource~\cite{kinjo2026togomcp}.  TogoMCP is our main baseline.
\textsc{Autoschema} is an independent grounding framework that discovers this
evidence from live endpoints instead of relying on a prewritten MIE file.

Our evaluation combines two kinds of questions.  BioASQ provides realistic
factoid, list, and yes or no questions, but it does not guarantee that every
answer exists in a given graph~\cite{krithara2023bioasq}.  Graphlet based
generation instead ties each question to graph evidence
\cite{jonker2026biographletqa}.  We use this principle for Resource Focused
and Multi Resource Biomedical KGQA, and use several years of BioASQ Task B
for Longitudinal Biomedical Semantic QA.  We also test transfer to the
tmQM-RDF chemistry graph~\cite{cibinel2026tmqmrdf}.

\section{Method}
\label{sec:method}

\subsection{Problem Setting and Framework Overview}
\label{sec:problem}

An agent cannot write a reliable SPARQL query until it knows where the answer
may be stored and how that source represents the concepts in the question.
This information is difficult to assume in a collection of independently
maintained graphs.  Names, identifiers, relations, and links between graphs
can differ even when the graphs describe the same domain.  \textsc{Autoschema}
addresses this problem by observing the endpoints when a question arrives and
returning only the evidence needed for that question.

Figure~\ref{fig:autoschema-overview} shows the framework and its role.  The
agent interprets the question, plans, writes SPARQL, and checks the result.
\textsc{Autoschema} sits between the agent and
the live knowledge graphs.  When the agent lacks schema knowledge, it calls a
grounding module and receives a compact view of the current endpoint.  This
separation makes the framework independent of a particular agent or execution
system.

\textsc{Autoschema} is implemented as a standalone MCP server.  Its four core
functions are asynchronous and accept a database key as input.  An endpoint
resolver maps this key to a SPARQL endpoint and its named graphs.  A deployment
can also pass an endpoint URL directly.  The core functions do not call a
language model.  They issue SPARQL queries, format the observed results as
grounding evidence, and return that evidence to the calling agent.  Query
generation and execution remain outside the framework.  This interface allows
the same grounding layer to be used with different agents and model families.

More formally, let $\mathcal{D}=\{G_1,\ldots,G_m\}$ be a set of RDF graphs and
let $q$ be a natural language question.  To answer $q$, the agent must ground
four decisions.  It must identify the relevant source, resolve question terms
to graph entities, find the relations and paths that express the request, and
connect sources when one graph is not enough.  The four modules in the center
of Figure~\ref{fig:autoschema-overview} support these decisions.

On first uncached access to a source, the framework builds a bounded live index and
caches it in memory and on disk.  The index records frequent classes,
properties, object kinds, and real example values.  Each source has one cache
entry.  Calls within a process use memory, and later runs can load a JSON copy
from disk.  Refresh is optional and can be requested at any time to rebuild the
index from the current endpoint.  Entity search and local path expansion are
not cached and still query the endpoint when called.

The index is a sampled view of the endpoint rather than a complete ontology.
This choice keeps onboarding bounded on large graphs while giving the agent
enough evidence to form and test a query.  Appendix~\ref{app:tool-interfaces}
gives the exact sampling limits and function signatures.

\begin{table}[t]
  \caption{Evaluation tasks.  Graph-supported means that each gold answer is
  verified by executing a gold SPARQL query.}
  \label{tab:task-overview}
  \centering
  \footnotesize
  \setlength{\tabcolsep}{3pt}
  \begin{tabularx}{\columnwidth}{@{}>{\raggedright\arraybackslash}p{0.34\columnwidth}X@{}}
    \toprule
    Task & Data and evaluation focus \\
    \midrule
    Resource Focused Biomedical KGQA & RDF Portal. Factoid and list questions within one resource. \\
    Multi Resource Biomedical KGQA & RDF Portal. Factoid and list questions across resources. \\
    Longitudinal Biomedical Semantic QA & Six BioASQ Task B sets. Factoid, list, and yes or no questions collected over time. \\
    Chemistry Knowledge Graph Transfer & tmQM-RDF. Factoid and list questions on a graph outside biomedicine. \\
    \bottomrule
  \end{tabularx}
\end{table}

\subsection{Live Grounding Modules}
\label{sec:autoschema-tools}

The modules follow a broad to focused process.  They first locate a useful
source and then reveal the specific entities and paths needed to form a query.
Each module has a distinct role and returns evidence at a different level of
detail.

\textbf{Source summarization} answers ``which graph should the agent query?''
The source catalog first retrieves candidate databases from titles,
descriptions, and categories.  The agent then calls
\texttt{get\_rdf\_\allowbreak schema}.  Its inputs are a database and an optional entity
class.  To build its index, the function counts instances grouped by
\texttt{rdf:type}, keeps frequent classes, and retrieves stable example IRIs.
For each retained class, it samples instances and counts their predicates.  It
then samples predicate objects to determine whether a value is a literal, an
IRI, or a blank node.  For IRI objects, it also checks their observed RDF
types.

The returned summary includes the required named graph scope, class counts,
example entity IRIs, full predicate IRIs, observed range kinds, and example
values.  It also includes a simple query made from a real entity and predicate.
The optional \texttt{entity\_class} argument narrows the property view after a
relevant class is known.  These outputs let the agent distinguish, for
example, a literal identifier from an IRI valued link before it writes a
filter or join.

\textbf{Entity resolution} answers ``how does this graph identify the thing
named in the question?''  It searches labels and common accession forms and
returns candidate graph identifiers.  Concretely,
\texttt{search\_rdf\_\allowbreak entity(database, keyword)} searches common label, title,
name, identifier, and alternative label predicates.  It also recognizes common
biomedical accession forms, such as OBO and UniProt identifiers, that may be
encoded directly in an IRI rather than stored as a label.

Only URI subjects are returned as anchors.  Blank node identifiers are not
returned because they are not stable across separate SPARQL requests.  Each
result pairs an IRI with the label or identifier that matched.  The agent uses
the question context and the source summary to choose among candidates.  This
step prevents a surface name from being copied into a query when the endpoint
uses a different identifier.

\textbf{Relation and path discovery} answers ``how are the requested entities
or values connected?''  Starting from a resolved entity, it examines nearby
relations and short paths.  The \texttt{expand\_\allowbreak entity} function takes a
database and a stable entity IRI.  It uses one SPARQL request to retrieve
direct properties and paths through up to three nested blank nodes.  A nested
path must be explored within one request because its internal blank node
identifiers cannot be reused later.

The response groups direct properties by predicate and shows a few observed
values.  For nested structures, it returns the predicate chain, its leaf
values, and an executable graph scoped SPARQL pattern.  Paths ending in likely
value fields such as labels, identifiers, and names are shown first.  The
agent can copy the returned pattern into its candidate query when a value must
be filtered or joined.  This local expansion provides a concrete query shape
that is often missing from a class level summary.

\textbf{Cross-database bridging} answers ``how can evidence from two graphs be
connected?''  The \texttt{get\_cross\_db\_\allowbreak bridges} function checks three
signals.  It first uses an optional identifier conversion service when one is
configured.  It then checks whether an IRI valued property in one source points
directly into an entity namespace observed in the other source.  Finally, it
looks for property names observed in both live indexes and ranks identifier
like properties above generic display fields.

Direct IRI reuse is the strongest live signal because the object in one graph
is already an entity in the other.  A shared property name is weaker because
the two properties may have different meanings.  The module therefore returns
bridge candidates and a SPARQL join skeleton rather than claiming that a join
is correct.  The agent must execute the skeleton and inspect its rows before
using the bridge in an answer.

Together, the modules avoid loading a complete schema before every question.
The agent can stop after source and entity grounding for a simple query, use
path discovery when the graph structure is unclear, and request a bridge only
when the question spans multiple sources.

\subsection{Agentic Query Construction}
\label{sec:agent-loop}

The left side of Figure~\ref{fig:autoschema-overview} shows how the modules are
used in an agent loop.  The agent first decides whether a knowledge graph is
appropriate for the question.  If so, it calls the modules needed for its
current subgoal, writes a candidate SPARQL query, and runs it on the selected
endpoint.  A useful result can support the final answer.  An empty result or
an endpoint error provides feedback that the selected entity, relation, path,
or source may be wrong.  The agent can then request more focused grounding
evidence and revise the query.

The loop is a policy rather than a fixed pipeline.  For a graph question, the
agent normally selects a source and requests its summary once.  It resolves
named entities before writing entity specific triples and calls local
expansion only when the required predicate is unclear.  The bridge module is
reserved for questions that genuinely require two sources.  Tool responses
are added to the model context in a compact textual form.  If two attempts at
the same subproblem fail, the evaluation policy asks the agent to change the
predicate, source, or grounding tool instead of repeating the same query.

This loop is important because no single schema view is sufficient for every
question.  Source summaries guide the first choice, entity and path evidence
support query construction, and execution results reveal what still needs to
be corrected.  The process ends when the agent has enough evidence to answer
or reaches the shared interaction limit described in
Section~\ref{sec:eval-protocol}.
Appendix~\ref{app:tool-interfaces} gives the concrete tool signatures,
sampling limits, and evaluation prompt.

\subsection{A Live Grounding Example}
\label{sec:grounding-example}

Consider the question in Figure~\ref{fig:case-study}: ``What value is
associated with the strain that shares an identifier with CCM 2634 and also
relates to entity 11225?''  The live calls below are abbreviated from a
successful BacDive trace.  IRI prefixes are shortened only for readability.

\begin{small}
\small
\noindent\textbf{1. Resolve the first question term.}\par\noindent
The call \texttt{search\_rdf\_\allowbreak entity} uses source \texttt{bacdive} and keyword
``CCM 2634.''  It returns collection record \texttt{20378} with the same
label.

\noindent\textbf{2. Call \texttt{expand\_entity}.}\par\nopagebreak[4]\noindent
For record \texttt{20378}, the response contains relation
\texttt{describesStrain}.  The observed object is \texttt{strain/7366}.

\noindent\textbf{3. Resolve the second question term.}\par\nopagebreak[4]\noindent
Searching \texttt{bacdive} for ``11225'' returns several candidates.  The
candidate with type \texttt{Strain} is \texttt{strain/11225}, labeled
``Nocardiopsis buerjinensis 11225.''

\noindent\textbf{4. Compare observed properties.}\par\nopagebreak[4]\noindent
Expanding the two strain IRIs returns \texttt{hasPhylum} $\rightarrow$
\texttt{Actinobacteria} for both entities.
\end{small}

The evidence grounds every graph specific part of the query.  In particular,
it shows that \texttt{CCM 2634} denotes a collection number record rather than
a strain, and that \texttt{describesStrain} connects that record to the strain.
It also identifies \texttt{hasPhylum} as the shared value relation.  The agent
can therefore form a candidate query with the two observed triple patterns
\texttt{strain:7366 bds:hasPhylum ?shared} and
\texttt{strain:11225 bds:hasPhylum ?shared}, scoped to the BacDive named graph.
Here, \texttt{strain:} and \texttt{bds:} abbreviate the observed BacDive
entity and schema namespaces.
The shared binding is \texttt{Actinobacteria}.  No bridge call is needed after
live resolution shows that both entities and the answer relation occur in the
same graph.  This example illustrates the main division of labor in
Figure~\ref{fig:autoschema-overview}.  The agent decides what must be compared,
while \textsc{autoschema} supplies the source, IRIs, and predicates observed at
the current endpoint.

\section{Experiments}
\label{sec:experiments}

\subsection{Research Questions}

We organize the evaluation around three questions.  \textbf{RQ1} asks whether
live grounding improves Resource Focused and Multi Resource Biomedical KGQA.
\textbf{RQ2} asks whether the gains hold in Longitudinal Biomedical Semantic
QA and Chemistry Knowledge Graph Transfer.
\textbf{RQ3} asks how live grounding affects interaction cost and which parts
of the framework account for its gains.

\subsection{Compared Frameworks}
\label{sec:grounding-conditions}

Our main comparison is between TogoMCP and
\textsc{autoschema} with live grounding.  We use the same models, questions,
endpoints, and interaction budgets in all settings:
\begin{description}
\item[No schema] provides database names and query execution but no schema
  evidence.  It serves as a lower bound.
\item[TogoMCP] provides an LLM assisted and manually reviewed MIE file with
  schema descriptions and worked queries for each database.
\item[\textsc{autoschema}] provides the live grounding modules in
  Section~\ref{sec:autoschema-tools} without an MIE file.
\item[Single shot] is an ablation of \textsc{autoschema} that allows at most
  one call to each grounding module per question.
\end{description}

\begin{table}[htbp]
  \caption{Resource Focused Biomedical KGQA with 559 questions.  Results are
  mean $\pm$ standard deviation over three runs.}
  \label{tab:resource-focused}
  \centering
  \small
  \setlength{\tabcolsep}{3pt}
  \begin{tabular}{llrrr}
    \toprule
    Model & Framework & Factoid acc. & List F1 & Limit rate \\
    \midrule
    \multirow{3}{*}{gpt-oss-120b}
      & No schema  & $0.164\pm.019$ & $0.122\pm.006$ & $53.3\pm1.8\%$ \\
      & TogoMCP    & $0.198\pm.009$ & $0.134\pm.001$ & $44.0\pm1.0\%$ \\
      & AutoSchema & $\mathbf{0.260\pm.005}$ & $\mathbf{0.199\pm.012}$ & $\mathbf{33.5\pm1.5\%}$ \\
    \midrule
    \multirow{3}{*}{gemma4-31b}
      & No schema  & $0.239\pm.022$ & $0.117\pm.009$ & $17.6\pm1.0\%$ \\
      & TogoMCP    & $0.258\pm.008$ & $0.134\pm.003$ & $20.2\pm0.8\%$ \\
      & AutoSchema & $\mathbf{0.281\pm.010}$ & $\mathbf{0.145\pm.006}$ & $17.1\pm2.3\%$ \\
    \bottomrule
  \end{tabular}
\end{table}

\subsection{Evaluation Tasks}
\label{sec:benchmarks}

Table~\ref{tab:task-overview} summarizes the four evaluation tasks introduced
in Section~\ref{sec:introduction}.  They move from questions with guaranteed
graph answers to open biomedical questions and finally to a graph outside the
development domain.

\textbf{Biomedical KGQA.} We adapt graphlet anchored generation to the RDF
Portal~\cite{jonker2026biographletqa}.  Each question begins with a connected
fragment sampled from real triples.  We reconstruct and execute its gold
SPARQL query, then ask a language model to express the fragment as a natural
language question without seeing the answer.  Resource Focused Biomedical
KGQA contains questions answered within one resource.  Multi Resource
Biomedical KGQA uses verified links across resources.

\textbf{Longitudinal Biomedical Semantic QA.} We use factoid, list, and yes or
no questions from six yearly BioASQ Task B test sets
~\cite{krithara2023bioasq}.  These open questions may or may not benefit from
an RDF source, so they also test whether the agent chooses graph tools when
appropriate.  Summary questions are outside our scope.

\noindent\textbf{Chemistry Knowledge Graph Transfer.}\par
We use tmQM-RDF
~\cite{cibinel2026tmqmrdf}, which has no pre-existing MIE and was not used to
develop the framework.  Section~\ref{sec:rq-external} describes its irregular
schema and the transfer protocol.

\subsection{Models, Metrics, and Protocol}
\label{sec:eval-protocol}

We evaluate two models with open weights, \textbf{gpt-oss-120b} and
\textbf{gemma4-31b}.  Each model is served locally through vLLM on one
NVIDIA A100 GPU.  We use the same task instructions and budget policy in all
settings.  The grounding part of the prompt directs each framework to its own
interface.  Appendix~\ref{app:agent-prompt} gives the \textsc{autoschema}
policy.  We use a limit of 20 iterations in every compared setting. We also report the \emph{limit
rate}, which is the fraction of questions that reach the final iteration.

Predictions are matched to gold synonyms using exact matching, lexical
matching, and an LLM judge for unresolved pairs.  We report the final
judge-tier score as top-1 accuracy for factoid questions, set F1 for list
questions, and accuracy for yes or no questions.  Unless noted otherwise,
each result is the mean $\pm$ standard deviation over three runs.

\subsection{Biomedical Knowledge Graph QA}
\label{sec:biomedical-kgqa}

Table~\ref{tab:resource-focused} reports Resource Focused Biomedical KGQA.
\textsc{Autoschema} gives the highest factoid accuracy and list F1 for both
models.  It also lowers the limit rate relative to TogoMCP.  The gains are
larger for gpt-oss-120b, which reaches the interaction limit much more often under
the other frameworks.

Table~\ref{tab:multi-resource} reports Multi Resource Biomedical KGQA.
\textsc{Autoschema} again gives the highest mean factoid accuracy for both
models and lowers the limit rate relative to TogoMCP.  Variation across runs
is larger than in the resource focused task, so we treat the accuracy gains as
directional rather than conclusive.  Trace analysis shows that many questions
are solved through shared reference values found by local entity expansion.
The dedicated bridge module is rarely selected, which we examine further in
Section~\ref{sec:ablations}.

\begin{table}[htbp]
  \caption{Multi Resource Biomedical KGQA with 226 questions.  Results are
  mean $\pm$ standard deviation over three runs.}
  \label{tab:multi-resource}
  \centering
  \small
  \setlength{\tabcolsep}{3pt}
  \begin{tabular}{llrrr}
    \toprule
    Model & Framework & Factoid acc. & List F1 & Limit rate \\
    \midrule
    \multirow{3}{*}{gpt-oss-120b}
      & No schema  & $0.178\pm.004$ & $0.138\pm.001$ & $45.7\pm1.4\%$ \\
      & TogoMCP    & $0.167\pm.024$ & $0.165\pm.005$ & $49.4\pm0.3\%$ \\
      & AutoSchema & $0.203\pm.020$ & $0.200\pm.019$ & $\mathbf{40.0\pm1.1\%}$ \\
    \midrule
    \multirow{3}{*}{gemma4-31b}
      & No schema  & $0.110\pm.027$ & $0.105\pm.010$ & $25.1\pm2.0\%$ \\
      & TogoMCP    & $0.140\pm.008$ & $0.125\pm.019$ & $32.9\pm1.6\%$ \\
      & AutoSchema & $\mathbf{0.216\pm.041}$ & $0.134\pm.003$ & $\mathbf{26.1\pm1.6\%}$ \\
    \bottomrule
  \end{tabular}
\end{table}

\subsection{Longitudinal Biomedical Semantic QA}
\label{sec:rq-bioasq}

Table~\ref{tab:bioasq-core} reports an early and a recent BioASQ test set for
both models.
Unlike the graph supported tasks, BioASQ also tests whether the agent avoids
unnecessary graph queries.  \textsc{Autoschema} improves mean factoid accuracy
over TogoMCP for both models in both years.  The gains are clearer on Task 8B
than on the more variable Task 13B.  List and yes or no results are mixed,
which suggests that the main benefit is more reliable retrieval of short
entity answers.

\begin{table}[htbp]
  \caption{Longitudinal Biomedical Semantic QA on two BioASQ Task B test sets.
  Results are mean $\pm$ standard deviation over three runs.}
  \label{tab:bioasq-core}
  \centering
  \small
  \setlength{\tabcolsep}{2.2pt}
  \begin{tabular}{llrrr}
    \toprule
    Model & Framework & Fact. & List F1 & Yes/no \\
    \midrule
    \multicolumn{5}{l}{\emph{8B (2020)}} \\
    \multirow{3}{*}{gpt-oss-120b} & No schema  & $0.468\pm.008$ & $\mathbf{0.404\pm.031}$ & $0.829\pm.026$ \\
                                  & TogoMCP    & $0.510\pm.024$ & $0.360\pm.027$ & $0.838\pm.015$ \\
                                  & AutoSchema & $\mathbf{0.581\pm.040}$ & $0.396\pm.022$ & $\mathbf{0.860\pm.014}$ \\
    \multirow{3}{*}{gemma4-31b}   & No schema  & $0.614\pm.027$ & $0.494\pm.013$ & $0.875\pm.013$ \\
                                  & TogoMCP    & $0.565\pm.030$ & $0.466\pm.008$ & $0.877\pm.004$ \\
                                  & AutoSchema & $\mathbf{0.625\pm.030}$ & $\mathbf{0.500\pm.014}$ & $\mathbf{0.897\pm.014}$ \\
    \midrule
    \multicolumn{5}{l}{\emph{13B (2025)}} \\
    \multirow{3}{*}{gpt-oss-120b} & No schema  & $0.611\pm.048$ & $0.334\pm.038$ & $\mathbf{0.862\pm.019}$ \\
                                  & TogoMCP    & $0.554\pm.060$ & $0.347\pm.043$ & $0.850\pm.014$ \\
                                  & AutoSchema & $0.632\pm.046$ & $\mathbf{0.384\pm.014}$ & $0.837\pm.031$ \\
    \multirow{3}{*}{gemma4-31b}   & No schema  & $0.660\pm.006$ & $\mathbf{0.503\pm.019}$ & $0.801\pm.019$ \\
                                  & TogoMCP    & $0.646\pm.027$ & $0.458\pm.006$ & $0.825\pm.019$ \\
                                  & AutoSchema & $\mathbf{0.681\pm.040}$ & $0.453\pm.012$ & $\mathbf{0.846\pm.025}$ \\
    \bottomrule
  \end{tabular}
\end{table}

\FloatBarrier

\subsection{Results Across BioASQ Task B Years}
\label{sec:rq-generalization}

\begin{figure*}[htbp]
  \centering
  \includegraphics[width=\textwidth]{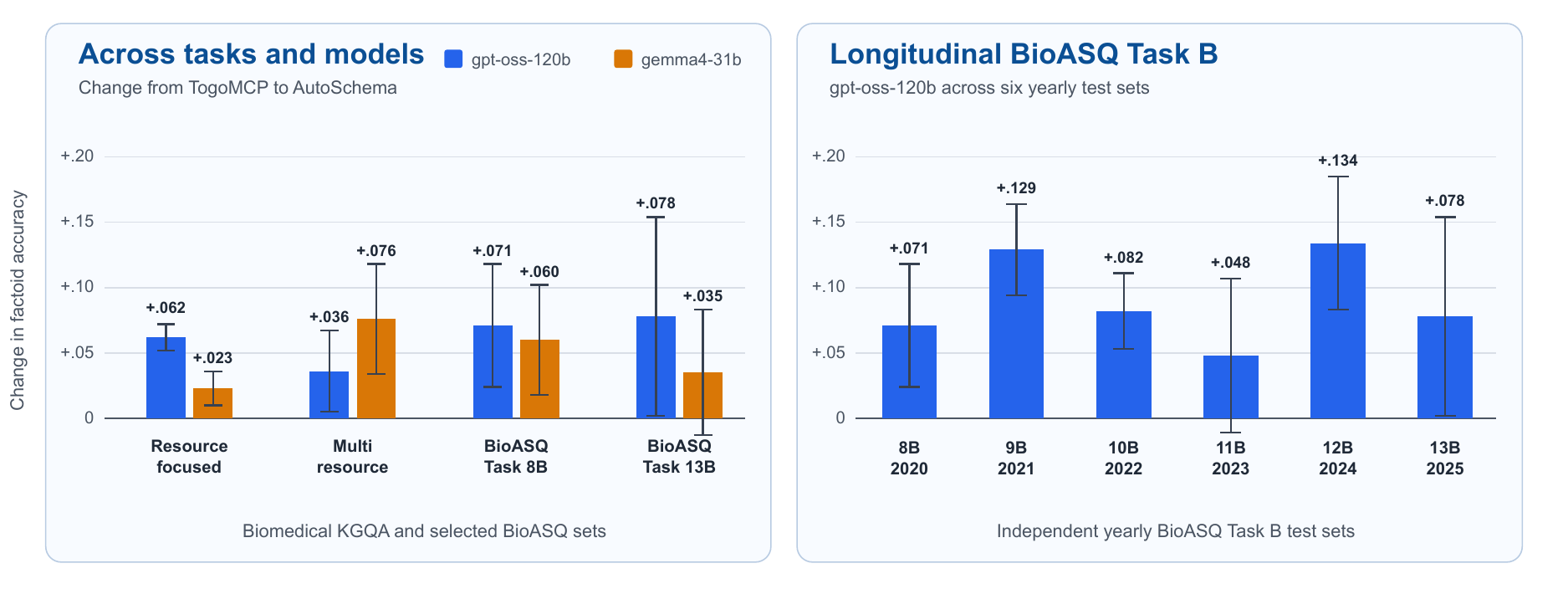}
  \caption{Factoid accuracy gain of \textsc{autoschema} over TogoMCP.
  Bars show differences between three-run means.  Error bars show the root sum
  square of the reported standard deviations and are descriptive rather than
  confidence intervals.}
  \Description{Two bar charts show the change in factoid accuracy from
  TogoMCP to AutoSchema.  The first compares two models across Resource
  Focused Biomedical KGQA, Multi Resource Biomedical KGQA, BioASQ Task 8B,
  and BioASQ Task 13B.  The second shows gpt-oss-120b results across six yearly
  BioASQ Task B test sets.  Every mean difference is positive.}
  \label{fig:experiment-summary}
\end{figure*}

We extend the gpt-oss-120b comparison to all six yearly BioASQ Task B test sets.
Figure~\ref{fig:experiment-summary} shows the change in mean factoid accuracy
from TogoMCP to \textsc{autoschema}.  The mean is higher in every year.  The
size of the gain varies, and run variation makes some individual years less
conclusive.  The consistent direction across independently collected test
sets provides stronger evidence of temporal robustness than any one year.

\begin{figure*}[htbp]
  \centering
  \includegraphics[width=0.99\textwidth]{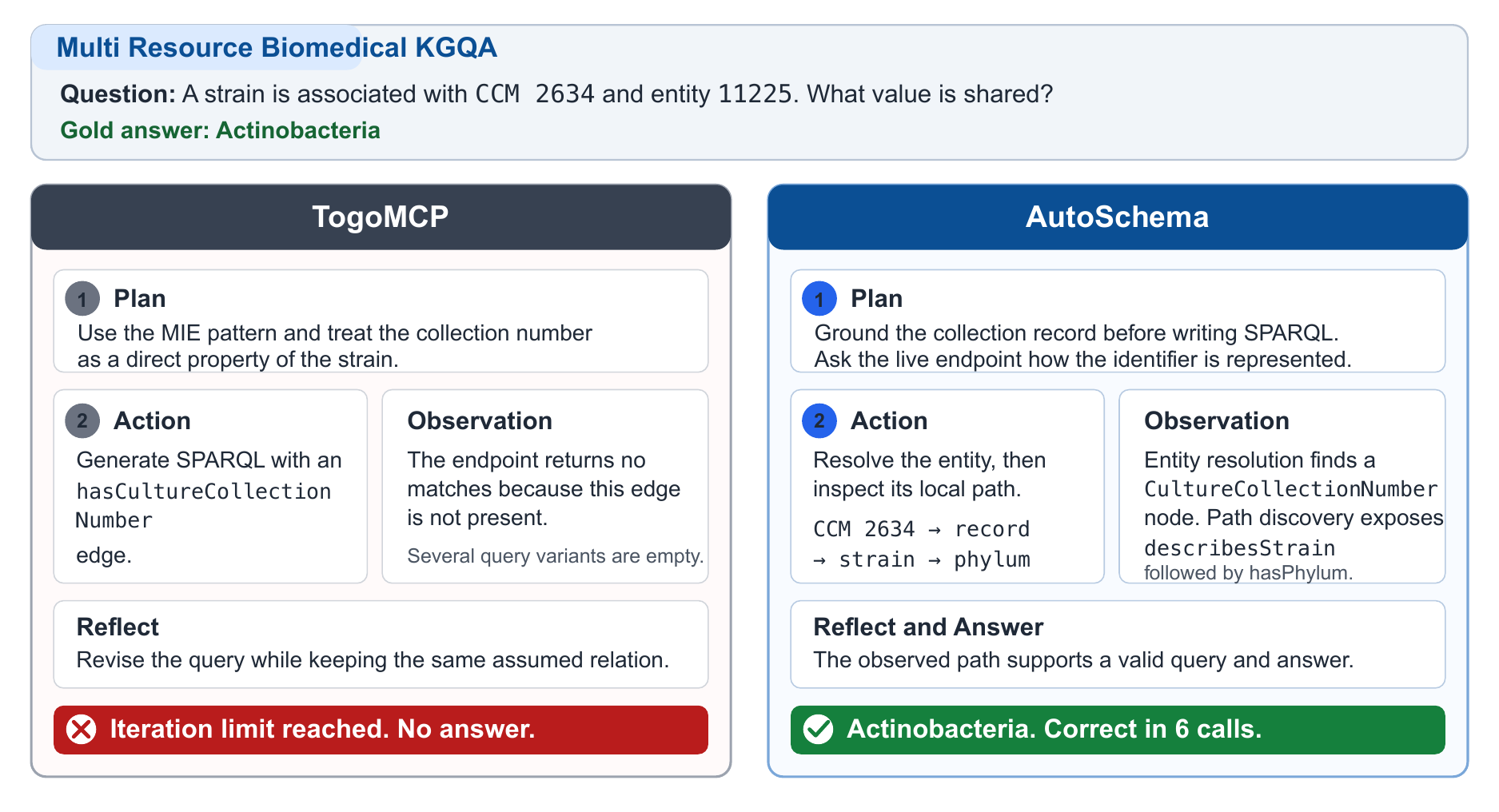}
  \caption{Paired trace for a Multi Resource Biomedical KGQA question.
  TogoMCP continues from an incorrect static relation and reaches the
  interaction limit.  \textsc{Autoschema} uses live entity resolution and
  local path discovery to return the correct answer.  The example explains a
  mechanism and does not estimate how often it occurs.}
  \Description{A question and gold answer appear above two agent traces.  The
  TogoMCP trace plans a direct collection-number relation, issues queries that
  return no matches, and reaches the iteration limit.  The AutoSchema trace
  resolves the collection record, discovers its path to a strain and phylum,
  and returns Actinobacteria in six calls.}
  \label{fig:case-study}
\end{figure*}

\subsection{Cost and Latency}
\label{sec:cost-latency}

Live grounding adds endpoint requests, so we measure both wall clock latency
and tool calls.  Table~\ref{tab:cost-latency} pools the two biomedical KGQA
tasks and the two BioASQ test sets in Table~\ref{tab:bioasq-core}.
\textsc{Autoschema} uses fewer tool calls than TogoMCP for both models while
latency remains comparable.  The latency ordering differs by model and is
more variable.  Trace inspection suggests that an early schema lookup can
replace several failed SPARQL attempts based on guessed predicates.

\begin{table}[htbp]
  \caption{Mean per-question interaction cost across the two biomedical KGQA
  tasks and two selected BioASQ sets.  Values are mean $\pm$ standard
  deviation over three runs.}
  \label{tab:cost-latency}
  \centering
  \small
  \setlength{\tabcolsep}{3pt}
  \begin{tabular}{llrr}
    \toprule
    Model & Framework & Latency (s) & Tool calls \\
    \midrule
    \multirow{3}{*}{gpt-oss-120b}
      & No schema  & $71.4\pm5.6$ & $13.22\pm.04$ \\
      & TogoMCP    & $67.1\pm5.9$ & $12.91\pm.05$ \\
      & AutoSchema & $\mathbf{59.5\pm11.9}$ & $\mathbf{11.13\pm.12}$ \\
    \midrule
    \multirow{3}{*}{gemma4-31b}
      & No schema  & $80.5\pm1.3$ & $9.31\pm.08$ \\
      & TogoMCP    & $\mathbf{74.2\pm1.6}$ & $9.22\pm.05$ \\
      & AutoSchema & $76.4\pm3.2$ & $\mathbf{8.27\pm.23}$ \\
    \bottomrule
  \end{tabular}
\end{table}

\subsection{Chemistry Knowledge Graph Transfer}
\label{sec:rq-external}

The core evaluation uses biomedical graphs.  We therefore test transfer to a
3{,}200-complex subset of tmQM-RDF~\cite{cibinel2026tmqmrdf}.  The graph was
not used to develop \textsc{autoschema} and has no pre-existing MIE.  Its
schema is irregular because values are represented through nested nodes and
the same property can have several reported sources.

We construct 39 graph supported questions that cover lookup, aggregation,
source selection, and multi-hop traversal.  For TogoMCP, we create and check
an MIE using its standard workflow.  This takes 28.2 minutes, 221{,}834 model
tokens, and 106 tool calls.  Live indexing takes 37.4 seconds and no model
tokens.  The experiment therefore measures both onboarding and answer quality.

Table~\ref{tab:tmqm} reports one run and should be read as preliminary.
\textsc{Autoschema} improves factoid accuracy and lowers the limit rate, but
uses more time and tool calls.  List F1 is unchanged.  On a separate flat
control graph, both frameworks reach the same factoid accuracy.  These results
suggest that live discovery is most useful for irregular schemas, but more
graphs and repeated runs are needed.

\begin{table}[htbp]
  \caption{Chemistry Knowledge Graph Transfer on 39 tmQM-RDF questions with
  gemma4-31b.  Results are from one run.}
  \label{tab:tmqm}
  \centering
  \footnotesize
  \setlength{\tabcolsep}{2.4pt}
  \begin{tabular}{lrrrrr}
    \toprule
    Framework & Fact. & List F1 & Limit & Time (s) & Calls \\
    \midrule
    TogoMCP    & $0.273$          & $0.167$ & $35.9\%$          & $\mathbf{1060.3}$ & $\mathbf{15.1}$ \\
    AutoSchema & $\mathbf{0.394}$ & $0.167$ & $\mathbf{23.1\%}$ & $1306.5$          & $16.7$ \\
    \bottomrule
  \end{tabular}
\end{table}

\subsection{Ablations}
\label{sec:ablations}

\textbf{Iterative grounding.} Table~\ref{tab:ablation-singleshot} compares full
\textsc{autoschema} with the single shot ablation.  On Resource Focused
Biomedical KGQA, limiting each module to one call causes only small changes
for both models.  Most of the benefit therefore comes from the first grounding
request.  Multi Resource Biomedical KGQA is less uniform.  Results remain
similar for gemma4-31b, but gpt-oss-120b loses list performance under the limit.  This
suggests that follow-up grounding can matter when the agent discovers a second
source during query construction.

\begin{table}[!t]
  \caption{Iterative and single shot grounding.  Values are mean $\pm$
  standard deviation over three runs.  R and M denote the Resource Focused
  and Multi Resource Biomedical KGQA tasks.}
  \label{tab:ablation-singleshot}
  \centering
  \scriptsize
  \setlength{\tabcolsep}{1.6pt}
  \begin{tabular}{llrrrr}
    \toprule
    Task & Access & Fact. & List F1 & Limit & Time (s) \\
    \midrule
    \multicolumn{6}{l}{\emph{gpt-oss-120b}} \\
    R & Iterative   & $0.260\pm.005$ & $0.199\pm.012$ & $33.5\pm1.5\%$ & $69.6\pm17.6$ \\
    R & Single shot & $0.246\pm.013$ & $0.185\pm.011$ & $35.5\pm1.0\%$ & $66.3\pm1.9$ \\
    M & Iterative   & $0.203\pm.020$ & $\mathbf{0.200\pm.018}$ & $40.0\pm1.1\%$ & $77.2\pm19.4$ \\
    M & Single shot & $0.185\pm.017$ & $0.155\pm.010$ & $41.9\pm2.9\%$ & $68.4\pm1.6$ \\
    \midrule
    \multicolumn{6}{l}{\emph{gemma4-31b}} \\
    R & Iterative   & $0.281\pm.010$ & $0.145\pm.005$ & $17.1\pm2.3\%$ & $100.0\pm6.2$ \\
    R & Single shot & $0.272\pm.006$ & $0.133\pm.021$ & $16.5\pm0.1\%$ & $97.1\pm3.3$ \\
    M & Iterative   & $0.216\pm.041$ & $0.134\pm.003$ & $26.1\pm1.6\%$ & $105.7\pm3.1$ \\
    M & Single shot & $0.227\pm.004$ & $0.125\pm.013$ & $24.0\pm1.6\%$ & $103.1\pm2.0$ \\
    \bottomrule
  \end{tabular}
\end{table}

\subsection{Qualitative Analysis}
\label{sec:case-studies}

Figure~\ref{fig:case-study} compares two traces for the same Multi Resource
Biomedical KGQA question.  TogoMCP follows a relation pattern that is absent
from the endpoint and reaches the interaction limit.  \textsc{Autoschema}
first resolves the collection record and then discovers the local path to the
phylum.  The observed path supports the correct answer without an explicit
bridge call.

The other traces show the boundary of this benefit.  In Chemistry Knowledge
Graph Transfer, live inspection recovers the named graph and nested property
path when the MIE namespaces do not match the loaded graph.  On the flat
control graph, both frameworks return the same answer in the same number of
calls.  Live grounding is therefore most useful when the endpoint
representation is uncertain or differs from the static description.

\section{Conclusion}

We introduced \emph{live schema grounding} as an alternative to requiring a
prewritten and manually reviewed schema file before an agent can query a
knowledge graph.  We implemented it in \textsc{autoschema}, a general
framework for agentic text-to-SPARQL.  Its four capabilities expose source structure, entity
candidates, local relation paths, and potential cross-database connections
only when the agent needs them.  This division of labor leaves interpretation
and query planning to the language model while grounding structural choices in
observations from the current endpoint.

Across two open-weight models, \textsc{autoschema} consistently improves mean
factoid accuracy over TogoMCP on Resource Focused Biomedical KGQA and
Multi Resource Biomedical KGQA.  It also improves gpt-oss-120b across six
yearly BioASQ Task B test sets and reduces average tool use in the biomedical
evaluation.  The tmQM-RDF transfer study provides
additional, preliminary evidence that the approach can expose a previously
undocumented graph without first producing a curated schema file.  The
ablations also qualify the result: one live grounding request captures most of
the benefit in Resource Focused Biomedical KGQA, whereas follow-up access can
matter in Multi Resource Biomedical KGQA.

These findings establish live endpoint evidence as a practical grounding
interface, but not yet as a complete solution to automatic federation.  The
dedicated bridge capability was rarely selected, some multi resource gains are
noisy, and the external transfer experiment contains only one run.  A natural
next step is therefore to couple live discovery with explicit source and
bridge planning, stronger answer verification, and evaluations over more
previously unseen graphs.  More broadly, shifting schema access from static
prompt context to observable, question-time evidence offers a path toward
knowledge-graph agents that remain useful as their underlying data changes.

\section*{Ethics and Privacy Statement}

This study uses public benchmark questions and publicly accessible RDF
endpoints.  It does not involve human participants or collect personal data.
AutoSchema sends question-dependent lookup queries to selected endpoints.
These requests may reveal the scientific terms in a question to endpoint
operators and may add load to public services.  Our experiments use public
questions without private user information.  Deployments that handle
sensitive questions should use access-controlled endpoints, avoid storing raw
questions, and apply caching and rate limits.

The framework may return incomplete or incorrect answers because endpoint
content, entity matches, or generated queries can be wrong.  This risk is
especially important for biomedical questions.  AutoSchema is intended for
research and information retrieval, not for clinical decision making.  Its
answers should be checked against the returned source evidence before they are
used in practice.  We use locally hosted models with open weights, which
avoids sending questions to a closed model API.  Local deployment does not
remove the need for secure storage, access control, and responsible use.
Users are also responsible for following the terms and licenses of the data
sources they query.

\bibliographystyle{ACM-Reference-Format}
\bibliography{sample-base}

\appendix

\section{AutoSchema Tool Interfaces}
\label{app:tool-interfaces}

\textsc{Autoschema} is implemented as a standalone MCP server.  Its core
interface consists of four asynchronous functions.  The functions only query
RDF endpoints and do not call a language model.  A deployment maps each
\texttt{database} argument to an endpoint and zero or more named graphs.  It
may instead accept an endpoint URL directly.  Query execution remains outside
the framework.  The agent uses the returned evidence to construct a query and
sends that query to the executor available in its environment.

\begin{table}[H]
  \caption{Core MCP tools exposed by \textsc{autoschema}.  Defaults are shown
  in the signatures.}
  \label{tab:autoschema-interface}
  \centering
  \footnotesize
  \begin{tabularx}{\columnwidth}{@{}>{\raggedright\arraybackslash}p{0.34\columnwidth}X@{}}
    \toprule
    Tool and inputs & Returned grounding evidence \\
    \midrule
    \texttt{get\_rdf\_schema}\newline
    \texttt{database}, optional \texttt{entity\_class=""} &
    Query scope, frequent classes with counts and example IRIs, observed
    properties with range kinds and values, and example SPARQL queries.  The
    optional class name narrows the property view. \\
    \texttt{search\_rdf\_entity}\newline
    \texttt{database}, \texttt{keyword}, optional \texttt{limit=10} &
    Candidate entity IRIs and labels found through common label and identifier
    properties.  For recognized biomedical accessions, the response may
    provide a directly constructed IRI instead of a label match. \\
    \texttt{expand\_entity}\newline
    \texttt{database}, \texttt{entity\_iri} &
    Direct properties and paths that pass through as many as three nested
    blank nodes.  Each nested path includes a SPARQL pattern that can be run
    again to obtain a filterable or joinable result. \\
    \texttt{get\_cross\_db\_bridges}\newline
    \texttt{database\_a}, \texttt{database\_b} &
    Candidate links based on an optional identifier conversion service,
    direct IRI reuse, or property names observed in both sources.  The response
    includes a join skeleton when a shared property is found. \\
    \bottomrule
  \end{tabularx}
\end{table}

\subsection{Live Index and Output Scope}

Section~\ref{sec:autoschema-tools} describes how the live index is built.  The
current implementation surveys at most 25 classes and 40 properties per
class.  The property survey uses up to 300 sampled instances of a class.
Requests are issued with bounded concurrency.  The index is cached in memory
and on disk so that later calls do not repeat the endpoint work.  A deployment
can rebuild the cache when the source changes.

This interface returns observed evidence rather than a complete ontology.
A class or property that is absent from the summary may still exist outside
the bounded sample.  Individual endpoint failures are skipped instead of
aborting the entire index build.  The tools therefore help the agent form and
test a query, but they do not prove that an unobserved relation is absent.

\subsection{Source Catalog Adapter}

The standalone server can work with one endpoint or a registry of named
endpoints.  Our evaluation uses many named biomedical sources.  Its adapter
also exposes three catalog calls.  They are
\texttt{find\_\allowbreak databases},
\texttt{list\_\allowbreak databases}, and
\texttt{list\_\allowbreak categories}.  These calls select a source before
the four core tools are used.  Existing source titles and descriptions are retained when
available.  For a source without a catalog record, the adapter derives a
short entry from class and property names in the live index.  This entry helps
source selection but does not provide query patterns or manually written
schema relations.

\subsection{Evaluation Prompt Policy}
\label{app:agent-prompt}

The evaluation wrapper converts the MCP input schemas into function calls and
returns textual tool results to the agent.  Each result is limited to 6,000
characters before it is added to the model context.  The AutoSchema system
prompt applies the following policy.

\begin{enumerate}
  \item First decide whether the question calls for structured graph data.
  General facts and literature questions do not open an RDF source only as a
  precaution.
  \item For a graph question, call \texttt{find\_databases} once.  Then call
  \texttt{get\_rdf\_schema} once for each selected source.
  \item Use \texttt{search\_rdf\_entity} to map a name in the question to an
  IRI.  Use \texttt{expand\_entity} when the needed value is not a direct
  property or may be stored behind blank nodes.
  \item For a question that truly needs two sources, call the bridge tool once
  before constructing the join.
  \item Execute the candidate SPARQL with the required named-graph scope.  If
  two attempts at the same subproblem fail or return no rows, change the
  predicate, source, or tool instead of repeating the same attempt.
  \item Return a concise answer when enough evidence has been found.  If the
  interaction budget is nearly exhausted, return the best supported answer
  available rather than ending without an answer.
\end{enumerate}

The task instructions and the failure-pivot rule are shared across compared
frameworks.  The grounding block above replaces instructions for reading a
prewritten schema file with calls to the live interfaces.

\end{document}